\documentclass{article}

\usepackage{arxiv}
\usepackage{graphicx}
\usepackage{subfigure}
\usepackage{pgfplots}
\usepackage{multirow}
\usepackage{tikz}
\usepackage{subcaption}
\usepackage{booktabs} 
\usepackage{multirow}
\usepackage{xcolor}
\usepackage{hyperref}
\usepackage{amssymb}
\usepackage{pifont}
\usepackage{marvosym}  
\usepackage{mathtools}
\usepackage{amsthm}
\usepackage[capitalize,noabbrev]{cleveref}

\theoremstyle{plain}

\theoremstyle{definition}

\theoremstyle{remark}

\begin{document}

\title{Dual Atrous Separable Convolution for Improving Agricultural Semantic Segmentation}

\author{
	Chee Mei Ling\\
	Electrical and Computer. Engg.\\
	Lakehead University\\
	Thunder Bay, ON P7B5E1 \\
	\texttt{cling@lakeheadu.ca} 
	\And
    Thangarajah Akilan \\
	Software Engg.\\
	Lakehead University\\
	Thunder Bay, ON P7B5E1 \\
	\texttt{takilan@lakeheadu.ca}
    \And
    Aparna Ravinda Phalke\\
	Applied Science\\
	University of Alabama\\
	Huntsville, AL 35899 \\
	\texttt{arp0028@uah.edu}
}

\maketitle








\begin{abstract}
Agricultural image semantic segmentation is a pivotal component of modern agriculture, facilitating accurate visual data analysis to improve crop management, optimize resource utilization, and boost overall productivity.
This study proposes an efficient image segmentation method for precision agriculture, focusing on accurately delineating farmland anomalies to support informed decision-making and proactive interventions.
A novel Dual Atrous Separable Convolution (\texttt{DAS-Conv}) module is integrated within the DeepLabV3-based segmentation framework. The \texttt{DAS-Conv} module is meticulously designed to achieve an optimal balance between dilation rates and padding size, thereby enhancing model performance without compromising efficiency. 
The study also incorporates a strategic skip connection from an optimal stage in the encoder to the decoder to bolster the model’s capacity to capture fine-grained spatial features.
Despite its lower computational complexity, the proposed model outperforms its baseline and achieves performance comparable to highly complex transformer-based state-of-the-art (SOTA) models on the Agriculture-Vision benchmark dataset. It achieves more than $66\%$ improvement in efficiency when considering the trade-off between model complexity and performance, compared to the SOTA model.

\end{abstract}


\section{Introduction}\label{Introduction}

\begin{figure}[!tp]
\vskip 0.2in
\begin{center}
{\includegraphics[trim={0.2cm, 11.2cm, 0.95cm, 0cm}, clip, width=0.8\columnwidth]{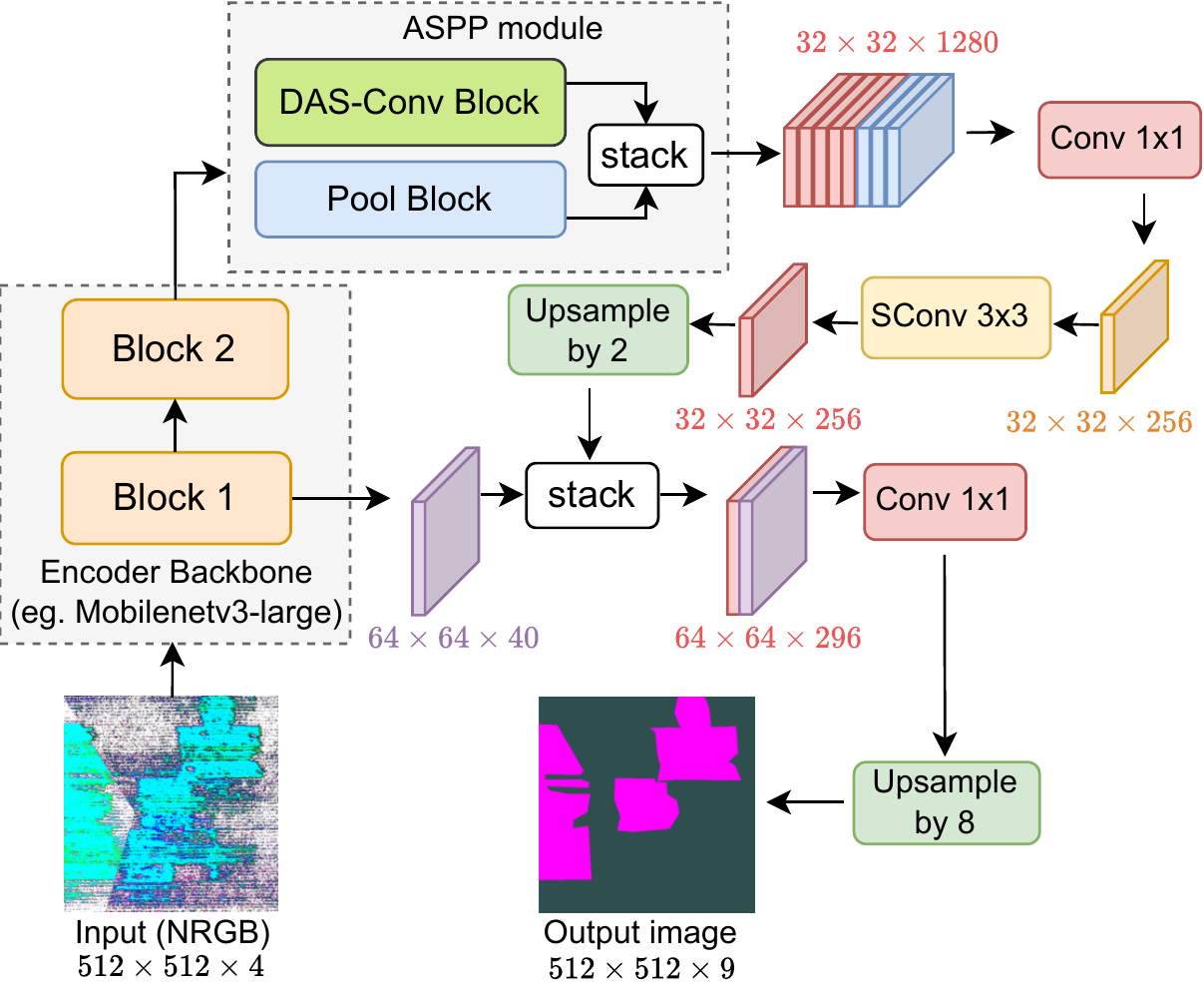} }
\caption{The proposed framework. For details, refer to \cref{table-backbone-structure}: encoder backbone, \cref{ASPP-table}: ASPP, and \cref{fig-DAS-conv}: \texttt{DAS-Conv}.}
\label{overall architecture}
\end{center}
\vskip -0.2in
\end{figure}


Over the years, advancements in artificial intelligence, computer technology, and machine learning (ML) have driven the rapid growth of computer vision (CV) in precision farming~\cite{binas2019reinforcement, ghazal2024computer}. CV-powered systems, often integrated with UAVs and advanced imaging sensors, have become foundational to precision agriculture by enabling real-time monitoring of crop health, soil conditions, nutrient levels, pest activity, and anomaly detection. These advancements significantly enhance productivity, efficiency, product quality, and sustainability~\cite{mizik2023can}. 

Meanwhile, semantic segmentation has gained significant attention within the CV and ML research communities due to its critical role in precise scene understanding and its applicability to various high-level downstream applications~\cite{suresh2025ecaseg, jahan2024improved, akilan2025self, wang2023np, 9994679, ozturk2024novel}.
In this direction, agricultural image semantic segmentation--a process that classifies every pixel in an image into meaningful categories, viz. crops, weeds, soil, and water--has emerged as a cornerstone of modern agriculture. It underpins critical high-level operations, including crop health assessment, yield estimation, irrigation optimization, and automation of overall agricultural operations, like supporting the navigation and functionality of agricultural robots and drones~\cite{ali2024ai, mendoza2024convolutional, luo2023semantic}.


Although highly complex vision transformer (ViT) models have shown strong performance in pixel-level tasks like segmentation~\cite{hatamizadeh2023global, jiageminifusion}, convolutional encoder-decoder (Conv-EnDec) architectures, such as UNet~\cite{ronneberger2015u} and DeepLab~\cite{chen2017rethinking}, remain the preferred choice for remote sensing applications. This preference is particularly evident in domains like precision agriculture, where computational efficiency and memory constraints are critical considerations.
In this context, we propose an enhanced semantic segmentation model based on DeepLabv3~\cite{chen2017rethinking} for detecting farmland anomalies in aerial agricultural imagery. The proposed approach aims to address the challenges of precision agriculture by balancing segmentation performance with computational efficiency.

The organization of this paper is as follows: \cref{Related-Work} reviews the relevant literature, providing context for this work. \cref{Methodology} outlines the proposed approach and its novelty. \cref{Experiments} describes the experimental setup, presents the results, and analyzes the model's performance. \cref{conclusion} summarizes the findings and offers ideas for future research.

\section{Related Work} \label{Related-Work}

\begin{table*}[!ht]
\caption{A succinct comparisons of key related works}\label{table-literature-review}
\begin{center}
\begin{small}
\setlength{\tabcolsep}{1pt} 
\renewcommand{\arraystretch}{1.3} 
\begin{tabular}{p{0.2cm} p{0.13\linewidth} p{0.18\linewidth} p{0.25\linewidth} p{0.4\linewidth}}
\toprule
 & \sc Ref & \sc Architecture & \sc Approach & \sc Limitations \\
\midrule
\multirow{5}{*}{\rotatebox{90}{\sc ViT Models}} 
 & Shen et al.~\cite{shen2022aaformer} & MiT-B3 encoder & Boundary maps concatenated with encoder output & Insignificant improvement, Suboptimal for datasets without boundary maps. \\ 

 & Tavera et al.~\cite{tavera2022augmentation} & SegFormer + MiT-B5 & Adaptive sampling and augmentation invariance methods & Similar absolute pixel counts for some categories reduce the effectiveness of adaptive sampling. \\ 

 & Yang et al.~\cite{yang2022agriculture} & SegFormer + MiT-B3 and MiT-B2 & Image mosaic scheme, ensemble modeling & Long training times due to heavier models and increased data generation. \\
\midrule

\multirow{6}{*}{\rotatebox{90}{\sc Conv-EnDec Models}} 
 & Yang et al.~\cite{yang2020reducing} & DeepLabv3+ \newline + IBN-Net backbone & Switchable normalization block and hybrid loss function & No strong evidence that IBN-Nets reduce feature divergence for better results. \\

 & Khan et al.~\cite{khan2023segmentation} & ResNet101 backbone & Feature fusion and context aggregation modules in the decoder & Long training times. Difficulty in extracting boundary patterns due to coarse annotations. \\

 & Liu et al.~\cite{liu2022multi} & PAFNet + dual \newline MobileNetv3 encoders & PAF and GFU modules & Manual selection of primary modality required to guide secondary modality learning. \\

 & Sheng et al.~\cite{sheng2020effective} & EfficientNet-B0/B2 \newline encoder + DeepLabV3 & GVI and AGN modules & Impact of vegetation index is unclear. Generalizability and all training strategies are untested. \\
\bottomrule
\end{tabular}
\end{small}
\end{center}
\vskip -0.1in
\end{table*}


The existing key studies are grouped into: ViT-based models and Conv-EnDec-based models as summarized in \cref{table-literature-review}.

\subsection{ViT-based Models}

Transformer architectures were originally introduced in natural language processing~\cite{vaswani2017attention}, where their ability to model global dependencies and handle sequential data efficiently revolutionized the field. Building on these strengths, researchers extended transformers to computer vision tasks, leading to the development of Vision Transformers (ViT)~\cite{dosovitskiy2021an}, which treat image patches as tokens. ViTs introduced a significant shift from the dominance of traditional convolutional neural networks (CNNs), particularly in scenarios involving large-scale datasets and computational resources. ViTs have shown that encoding images as a series of embedded patches and using the self-attention mechanism to capture contextual information is effective. Subsequent research built upon these ideas and proposed various improvements to ViTs. For instance, Shen~\textit{et~al.}~\cite{shen2022aaformer} introduced a multi-modal transformer network, named AAFormer, for farmland anomaly recognition. This model integrates mix-transformer (MiT) and squeeze-and-excitation (SE) modules to improve feature extraction. Additionally, it exploits boundary features at the decoder stage to improve performance. However, its reliance on additional features may limit its applicability in scenarios where such information is unavailable, highlighting a broader challenge in adapting it to domains with limited labeled data.

Tavera~\textit{et~al.}~\cite{tavera2022augmentation} proposed an adaptive sampling approach that selects training images based on pixel distribution and the network's confidence scores. They evaluated their method using a SegFormer~\cite{xie2021segformer} architecture with a MiT-B5 backbone, focusing on semantic segmentation of agricultural aerial images. While their adaptive sampling technique improved performance for certain classes, their training configuration for the dataset revealed limitations, particularly when addressing class imbalance. This issue led to a modest drop in performance for categories with small differences in absolute pixel counts.

Yang~\textit{et~al.}~\cite{yang2022agriculture} came up with a method combining pretrained ViTs, mosaic downsampling, and ensemble learning for agricultural pattern recognition. ViTs captured global contextual features, mosaic downsampling processed high-resolution aerial images, and ensemble learning improved robustness by aggregating predictions. While effective in handling multimodal data (RGB + NIR), it is computationally intensive, may obscure fine-grained details, and introduces increased complexity, making it less suitable for resource-constrained or real-time applications.



\subsection{Conv-EnDec-based Models}


In addition to advancements in ViT-based models, the Conv-EnDec architectures have made substantial progress in image segmentation due to their effectiveness in learning hierarchical features. 
Through a series of convolutional (Conv) layers, these models extract low-level features (e.g., edges and textures) in early layers and progressively learn higher-level abstractions (e.g., shapes and objects) in deeper layers, capturing spatial and contextual information effectively. 
This capability has driven their widespread adoption in several precision agriculture applications, even under challenging conditions such as variations in lighting, crop types, or environmental factors~\cite{gao2024accurate}.

For instance, Ulku~\cite{ulku2024reslmffnet} developed a specialized architecture, ResLMFFNet, to improve semantic segmentation accuracy while maintaining efficient inference speed, specifically targeting real-time precision agriculture applications. 
Khan~\textit{et~al.}~\cite{khan2023segmentation} proposed a deep CNN with feature fusion and context aggregation modules designed to address challenges in distinguishing agricultural patterns with similar visual features, which often lead to pixel misclassification. In the context of multimodal agricultural aerial images, such as RGB and NIR, Yang~\textit{et~al.}~\cite{yang2020reducing} highlighted that naively combining modalities without accounting for feature divergence can result in suboptimal performance. To mitigate this issue, they incorporated a switchable normalization block and a hybrid loss function into their model, effectively reducing feature divergence and improving segmentation accuracy.
Hence, Liu~\textit{et~al.}~\cite{liu2022multi} introduced a multimodality network that subsumes a pyramidal attention fusion (PAF) module, which captures fine-grained contextual representations of each modality using cross-level and cross-view attention fusion mechanisms. 
In addition, they used a gated fusion unit (GFU) to integrate complementary information from multiple modalities. Sheng~\textit{et~al.}~\cite{sheng2020effective} developed a lightweight generalized vegetation index (GVI) module to incorporate vegetation index information as an auxiliary input. 
To train models with GVI data, they also employed an additive group normalization (AGN) module that stabilizes the training and improves the overall performance. 
 
In summary, both ViT-based models and Conv-EnDec architectures are effective for tasks like semantic segmentation. However, in remote sensing applications on UAVs with limited memory and computational power, lightweight models are preferred. ViTs, with their complex self-attention mechanisms, are often too resource-intensive for such environments. In contrast, Conv-EnDec models are more efficient and can be optimized for deployment on edge devices, providing a better balance between performance and resource usage for real-time image analysis. However, there is still room for further research to enhance their effectiveness. Therefore, this work is timely and presents an efficient approach to improving the results of semantic segmentation.

\section{Methodology}\label{Methodology}


\cref{overall architecture} is a high-level overview of the proposed framework. It utilizes a Conv-EnDec, in which, we introduce \texttt{DAS-Conv} module into the decoding path to enhance the model performance. 
The overall model comprises $\approx 7.59$M parameters and requires $6.32$G multiplication-addition operations (or GFLOPs).
The following subsections detail the individual subsystems of the proposed model.



\subsection{Encoder} \label{Backbone}
The vision encoder backbone is an {ImageNet pre-trained} MobileNetV3-Large~\cite{Howard_2019_ICCV}. 
It is a lightweight, yet powerful CNN optimized by the platform-aware network architecture search algorithm aka NetAdapt~\cite{yang2018netadapt}. 
It strikes a balance between computational efficiency and feature extraction capabilities, making it well-suited for real-time applications and resource-constrained environments, such as drone-based aerial imaging and edge computing platforms.
\cref{table-backbone-structure} details this backbone, subsuming two main blocks of Conv layers and bottleneck modules. 
The backbone contains $\approx 2.975$M trainable parameters and requires $1.84$ GFLOPs, ensuring both computational efficiency and strong performance.
In our implementation, we excluded the last three modules of the original MobileNetV3-Large, which consist of a $7\times7$ pooling layer and two $1\times1$ convolutional layers. Additionally, we added a skip connection from the first block to the decoder, reintroducing low-level features to enhance the semantic cues learned by the middle layers of the decoder.

\begin{table}[!tp]
\setlength{\tabcolsep}{10pt} 
\caption{Details of the Mobilenetv3-large encoder backbone. EXP - expansion size, OUT - the number of output channels, SE - if Squeeze-And-Excite block is used, NL - the type of nonlinearity used, S - stride, and BNECK - bottleneck block.}
\label{table-backbone-structure}
\begin{center}
\begin{small}
\begin{sc}
\setlength{\tabcolsep}{10pt} 
\begin{tabular}{cccccccccc}
\toprule 
 & Input & Module & Exp & Out & SE    & NL & S \\
\midrule
\multirow{8}{*}{\rotatebox{90}{Block 1}} & $512^2\times4$ & Conv, $3\times3$  & - & 16  & \ding{55}  & h-Swish & 2 \\
& $256^2\times16$           & Bneck, $3\times3$ & 16       & 16  & \ding{55}        & ReLU & 1 \\
& $256^2\times16$           & Bneck, $3\times3$ & 64       & 24  & \ding{55}         & ReLU & 2 \\
& $128^2\times24$           & Bneck, $3\times3$ & 72       & 24  & \ding{55}         & ReLU & 1 \\
& $128^2\times24$           & Bneck, $5\times5$ & 72       & 40  & \ding{52}  & ReLU & 2 \\
& $64^2\times40$            & Bneck, $5\times5$ & 120      & 40  & \ding{52}  & ReLU & 1 \\
& $64^2\times40$            & Bneck, $5\times5$ & 120      & 40  & \ding{52}  & ReLU & 1 \\
\midrule
\multirow{10}{*}{\rotatebox{90}{Block 2}} & $64^2\times40$            & Bneck, $3\times3$ & 240      & 80  & \ding{55}         & h-Swish & 2 \\
 & $32^2\times80$ & Bneck, $3\times3$ & 200 & 80  & \ding{55} & h-Swish & 1 \\
& $32^2\times80$            & Bneck, $3\times3$ & 184      & 80  & \ding{55}          & h-Swish & 1 \\
& $32^2\times80$            & Bneck, $3\times3$ & 184      & 80  & \ding{55}           & h-Swish & 1 \\
& $32^2\times80$            & Bneck, $3\times3$ & 480      & 112 & \ding{52}  & h-Swish & 1 \\
& $32^2\times112$           & Bneck, $3\times3$ & 672      & 112 & \ding{52}  & h-Swish & 1 \\
& $32^2\times112$           & Bneck, $5\times5$ & 672      & 160 & \ding{52}  & h-Swish & 2 \\
& $32^2\times160$           & Bneck, $5\times5$ & 960      & 160 & \ding{52}  & h-Swish & 1 \\
& $32^2\times160$           & Bneck, $5\times5$ & 960      & 160 & \ding{52}  & h-Swish & 1 \\
& $32^2\times160$           & Conv, $1\times1$  & -        & 960 & \ding{55}          & h-Swish & 1 \\
\midrule
\multicolumn{8}{c}{Total number of trainable parameters: 2.975 M} \\
\bottomrule
\end{tabular}
\end{sc}
\end{small}
\end{center}
\end{table}





\subsubsection{Skip connection}


In Conv-EnDec models, the strategic placement of skip connections between the encoder and decoder is key to balancing fine-grained spatial detail preservation and computational efficiency, thereby enhancing model performance. This study conducted extensive experiments to identify the optimal placement, evaluating its impact at various stages within the architecture. The results show that the best placement is between Block 1 of the encoder and the decoder, as shown in \cref{overall architecture}. 

\subsection{Decoder}\label{decoder}




The decoder, a COCO pre-trained DeepLabV3, is enhanced by integrating our \texttt{DAS-Conv} module (cf.~\cref{sec-dasconv}). 
This integration significantly improves the network's ability to capture fine-grained semantic details and boost overall segmentation performance. 
The enhanced decoder, as depicted in \cref{overall architecture}, processes feature maps generated by the encoder using an Atrous Spatial Pyramid Pooling (ASPP). These high-level features are combined with low-level details from the encoder via skip connections, resulting in a more comprehensive representation.
Finally, the combined features are upsampled $8\times$ to match the input's spatial dimensions, producing semantic segmentation predictions.

\subsubsection{ASPP Sub-network}

\begin{table}[!tp]
\caption{Layer-wise detail of the \texttt{DAS-Conv} integrated ASPP module. DR - dilation rate, and $//$ - parallel pathway indexes.}
\label{ASPP-table}
\vskip 0.15in
\begin{center}
\begin{small}
\begin{sc}
\setlength{\tabcolsep}{10pt} 
\begin{tabular}{lccccr}
\toprule
   & $//$ & Input & Module & DR & Output \\
\midrule
\multirow{5}{*}{\rotatebox{90}{DAS-Conv}} & $P_1$  & $32^2\times960$ & Conv, $1\times1$  & - & $32^2\times256$ \\
 & $P_2$ & $32^2\times960$ & DASConv, $3\times3$ & 4 & $32^2\times192$ \\
  & $P_3$ & $32^2\times960$ & DASConv, $3\times3$ & 8 & $32^2\times192$ \\
  & $P_4$ & $32^2\times960$ & DASConv, $3\times3$ & 12 & $32^2\times192$ \\
  & $P_5$ & $32^2\times960$ & DASConv, $3\times3$ & 24 & $32^2\times192$ \\
\midrule
\multirow{3}{*}{\rotatebox{90}{POOL}} & $P_{6}$ & $32^2\times960$ & AvgPool $1\times1$ & - & $1^2\times960$ \\
   &                                       & $1^2\times960$ & Conv $1\times1$ & - & $1^2\times256$ \\
   &                                       & $1^2\times256$ & Bilinear & - & $32^2\times256$ \\
\midrule
\multicolumn{6}{c}{Total number of trainable parameters: 4.61 M} \\
\bottomrule
\end{tabular}
\end{sc}
\end{small}
\end{center}
\vskip -0.1in
\end{table}

The ASPP sub-network is refined by replacing the original atrous Conv with our \texttt{DAS-Conv} module, as summarized in \cref{ASPP-table}.
Through extensive ablation study, we determined that the optimal dilation rates for the \texttt{DAS-Conv} modules are 4, 8, 12, and 24. The study revealed that these rates produce superior segmentation results, particularly due to the smaller feature maps ($32\times32$) generated by the encoder. It is worth noting that larger dilation rates introduce redundancy and fail to capture meaningful context.

This sub-network subsumes six parallel connections, $P_1$ - $P_6$, grouped into two blocks: \texttt{DAS-Conv} and Pooling.
In the \texttt{DAS-Conv} block, a $1\times1$ Conv layer captures local information w/o dilation, while the $3\times3$ \texttt{DAS-Conv} layers, w/t multiple dilation rates, extract multiscale features to enhance semantic understanding.
In the Pooling block, a single parallel pathway is employed. It begins with global average pooling, which computes the mean across the entire spatial dimensions, reducing each feature map from the encoder to a single vector. A standard Conv is then applied to reduce the number of channels, followed by a bilinear interpolation to upsample the feature maps to a desired size.

\begin{figure}[!tp]
\begin{center}
\centerline{\includegraphics[trim={0.00cm, 0.00cm, 0.00cm, 0.0cm}, clip, width=0.7\columnwidth]{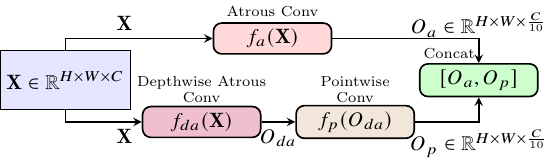}}
\caption{An illustration of the proposed \texttt{DAS-Conv} module.}
\label{fig-DAS-conv}
\end{center}
\vskip -0.2in
\end{figure}

\subsubsection{The DAS-Conv Module}\label{sec-dasconv}

\begin{figure}[!tp]
\begin{center}
\centerline{\includegraphics[trim={0.00cm, 13.00cm, 0.10cm, 0.0cm}, clip, width=0.7\columnwidth]{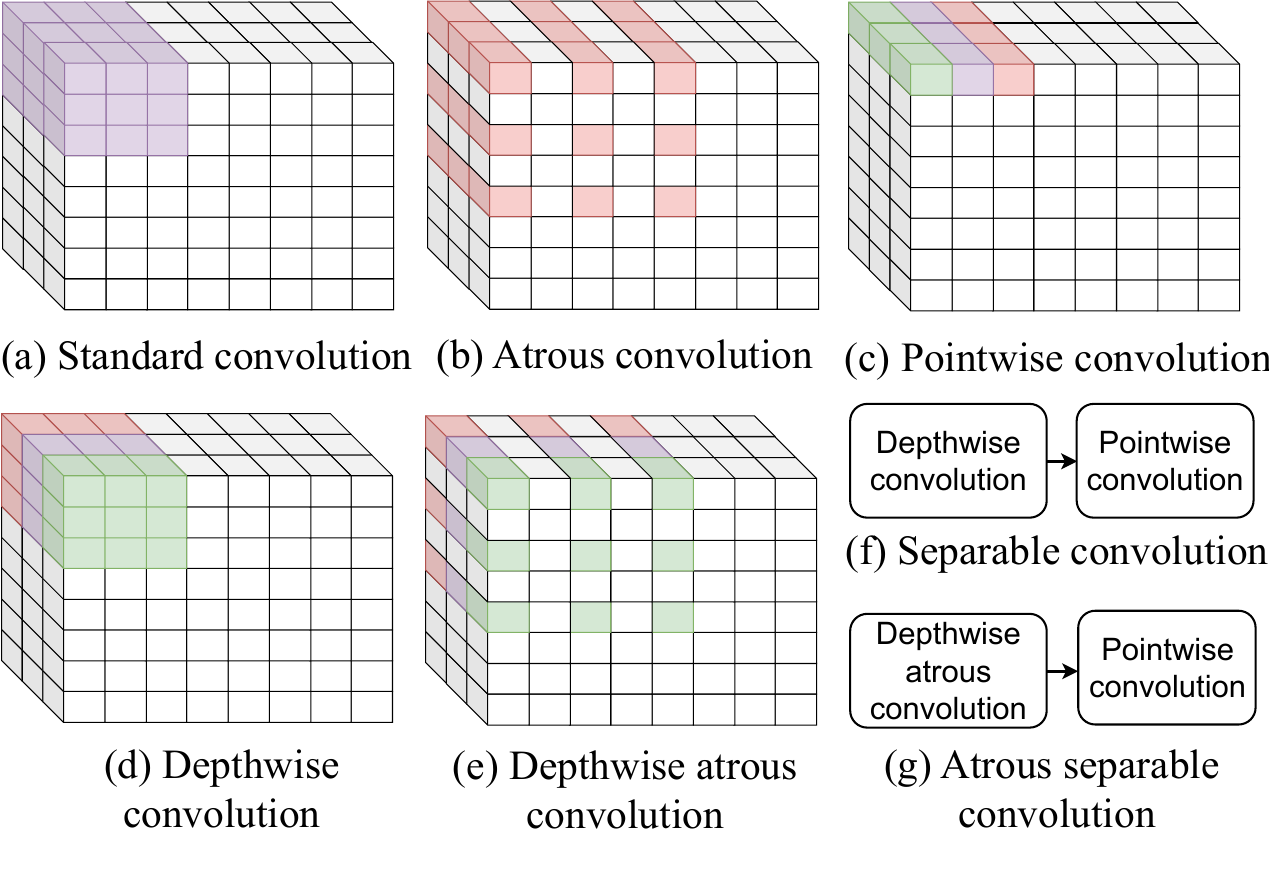}}
\caption{View of receptive field coverage for different Conv types and operation sequences in separable and atrous separable Conv.}
\label{fig-convolution-types}
\end{center}
\end{figure}

\cref{fig-DAS-conv} depicts the proposed \texttt{DAS-Conv} module, consisting of four different operations: two parallel operations---an atrous Conv $f_a(\cdot)$ parallel with sequentially connected depthwise atrous Conv $f_{da}(\cdot)$  and a pointwise Conv $f_p(\cdot)$---followed by a channel-wise concatenation. 
In the 1st parallel path, for an input $\mathbf{X} \in \mathbb{R}^{H \times W \times C}$, $f_a(\cdot)$ applies an atrous Conv with a kernel $K$ of size $3 \times 3$, a dilation rate $d$, and a stride $s$, to learn the spatial cues across all channels as:
\begin{equation} \label{eq-fa}
f_{a}(\mathbf{X}_{i,j}) = \sum_{m=0}^{M-1} \sum_{n=0}^{N-1} K(m,n) * \mathbf{X}(i \cdot s + m \cdot d, j \cdot s + n \cdot d),
\end{equation}
where $(i,j)$ are the coordinates of the input pixel, and $(m,n)$ are the coordinates of the kernel with dimensions $M \times N$.
\cref{fig-convolution-types}(b) conceptualizes the placement of an atrous Conv kernel of size $3\times3$ with $d=2$. 

In the 2nd parallel path, a depthwise atrous Conv $f_{da}(\cdot)$ is applied to the same input $\mathbf{X}$. It processes the input channel-wise using separate kernels for each channel as: 
\begin{equation} \label{eq-fda}
f_{da}(\mathbf{X}_{i,j}^{c}) = \sum_{m=0}^{M-1} \sum_{n=0}^{N-1} K^{c}(m,n) * \mathbf{X}^{c}(i \cdot s + m \cdot d, j \cdot s + n \cdot d),
\end{equation}
where $\mathbf{X}^{c}$ is the input feature map for the $c$-th channel, $K^{c}$ is the corresponding depthwise kernel for the $c$-th channel, and the rest of the terms are defined as in \cref{eq-fa}.
\cref{fig-convolution-types}(e) illustrates the placement of a depthwise atrous Conv kernel of size $3 \times 3$ with $d=2$, where the color coding represents separate kernels applied to individual channels. 
The output of $f_{da}(\cdot)$, then passes through a pointwise $1\times1$ Conv $f_p(\cdot)$. As depicted in \cref{fig-convolution-types}(c), this operation learns features across all channels using \cref{eq-fp}.
\begin{equation} \label{eq-fp}
f_p(i,j) = \sum_{c=0}^{C_{in}-1} K_p(c) * \mathbf{I}(i, j),
\end{equation}
where $(i,j)$ denotes output pixel coordinates, $C_{in}$ represents the number of input channels, $K_p$ denotes the $1 \times 1$ kernel at the $c$-th channel, and $\mathbf{I}$ represents the input.
To maintain the same spatial dimensions as the input, padding is applied as follows: $\left\lceil {d \cdot (K-1)}/{2} \right\rceil$,
where $K$ is the kernel size, and $d$ is the dilation rate used in $f_a(\cdot)$ and $f_{da}(\cdot)$. 
 Both $f_a(\cdot)$ and $f_p(\cdot)$ operations reduce the input channels to one-tenth while preserving the spatial dimensions, yielding outputs $f_a(\mathbf{X})\in\mathbb{R}^{32\times32\times96}$ and $f_p(f_{da}({X}))\in\mathbb{R}^{32\times32\times96}$.
These outputs are then concatenated channel-wise, generating the \texttt{DAS-Conv} output $\in\mathbb{R}^{32\times32\times192}$, as detailed in \cref{ASPP-table}. 



In summary, \texttt{DAS-Conv} combines the advantages of atrous and atrous separable convolutions. Atrous Conv enhances spatial learning by using dilated kernels that expand the receptive field without significant computational overhead. Atrous separable Conv, on the other hand, captures channel-wise dependencies via depthwise Conv followed by pointwise Conv. By integrating these operations, DAS-Conv simultaneously captures global and local spatial dependencies, both across and within channels, efficiently increasing the receptive field and improving feature learning.

\subsubsection{The Segmentation Prediction Head}



The outputs from all parallel connections in the ASPP sub-network are concatenated, resulting in a feature map of dimensions $32 \times 32 \times 1280$. This feature map is processed through a $1\times1$ Conv layer, reducing the number of channels to 256, followed by a dropout layer with a rate of 0.5. The reduced feature map is then passed through a separable Conv layer and upsampled by a factor of 2 using bilinear interpolation, resulting in a resolution of $64 \times 64 \times 256$.
We use a standard Conv instead of a separable Conv immediately after the ASPP output concatenation because standard convolutions effectively capture cross-channel interactions, combining both spatial and channel-level information more effectively. This enables the model to learn richer and more comprehensive feature representations.

The upsampled feature map with dimensions $64 \times 64 \times 256$ is concatenated with low-level features from Block 1 of the encoder backbone via a skip connection, which reintroduces fine-grained spatial details such as edges and boundaries. A $1\times1$ Conv layer then processes the concatenated feature map, reducing the number of channels to match the number of target classes.
Finally, the feature map is upsampled by a factor of 8 to match the spatial dimensions of the ground-truth labels, producing an output of dimensions $512 \times 512 \times 9$. To obtain the predicted segmentation map, softmax is applied along the channel dimension, followed by argmax to determine the most likely class for each pixel.

Note that, in this work, all Conv operations in the decoder are followed by Batch Normalization and ReLU activation function, to ensure feature refinement and avoid overfitting. 

\section{Experiments}\label{Experiments}

\subsection{Dataset}\label{Datasets}

This study uses the Agriculture-Vision Ver. 2 dataset~\cite{chiu2020agriculture} for model development and testing.
It contains high-resolution  
94,986 multispectral--RGB and near-infrared (NIR)--aerial images of size $512\times512$ with 9 anomaly annotations: double plant, drydown, endrow, nutrient deficiency, planter skip, storm damage, water, waterway, and weed cluster. Due to the scarcity of the storm damage category, it is excluded from the evaluation. The data set has samples in training, validation, and testing sets in a ratio of 6:2:2. Unlike typical aerial datasets, Agriculture-Vision images have a high resolution (10cm/pixel) with irregularly shaped anomaly regions and imbalanced class distributions as shown in \cref{fig-imbalance-dataset}. 
\subsubsection{Data Preprocessing}

\textbf{Label flattening:} The ground truth annotations in the dataset are non-exclusive, i.e., multiple labels can overlap within the same pixel. To simplify this, label images are flattened by prioritizing classes based on pixel frequency.
\textbf{Multi-spectral input:} To utilize both RGB and NIR details, we merge them as a 4-channel NRGB input data. \\
\textbf{Intensity normalization:} The input intensity values are rescaled to $[0.0, 1.0]$ by dividing them by 255.0 and then normalized channel-wise using mean of $[0.485, 0.456, 0.406]$ and standard deviation (STD) of $[0.229, 0.224, 0.225]$ for RGB, and mean of $0.485$ and STD of $0.229$ for NIR.


\begin{figure}[!t]
    \centering
    \begin{tikzpicture}        
        \begin{axis}[
            ybar,
            ymin=0,
            ymax=2.8,
            bar width=0.5cm, 
            xtick=data,
            xticklabels={\texttt{DD}, \texttt{ND}, \texttt{WC}, \texttt{WA}, \texttt{ER}, \texttt{DP}, \texttt{WW}, \texttt{PS}}, 
            ylabel={\footnotesize Total \# of pixels ($\times10^9$)}, 
            ylabel style={yshift=-0.5cm}, 
            width=1.0\columnwidth, 
            height=5cm, 
            xticklabel style={rotate=0, anchor=north}, 
            ymajorgrids, 
            grid style={gray!50}, 
            yminorgrids, 
            minor y tick num=4, 
            minor grid style={dashed, gray!50}, 
            minor tick style={gray!50}, 
        ]
            \addplot[fill=green!20, draw=green] coordinates { 
                (1,2.51)
                (2,1.45)
                (3,1.08)
                (4,0.224)
                (5,0.191)
                (6,0.167)
                (7,0.129)
                (8,0.0391)

            };
        \end{axis}         
    \end{tikzpicture} 
    \begin{flushleft}
    \scriptsize
    \noindent{\textbf{Legnd:}  \texttt{DD} - Dry Down, \texttt{ND} - Nutrient Deficiency, \texttt{WC} - Weed Cluster, \texttt{WA} - Water, \texttt{ER} - End Row,   \texttt{DP} - Double Plant, \texttt{WW} - Waterway, \texttt{PS} - Planter Skip }\vspace{-0.2cm}
    \end{flushleft}
    \caption{Semantic class vs pixel distribution in the dataset.}    \label{fig-imbalance-dataset}    \vspace{-0.2cm}
\end{figure}

\subsection{Implementation Details}


All experiments in this study were conducted on an A100-40GB GPU within the Alliance Canada computing infrastructure (Narval cluster), and the PyTorch Ver. 2.4.0 was used for model development.  The models were trained using the stochastic gradient descent (SGD) optimizer, with an initial learning rate of 0.001, momentum of 0.9, and a weight decay factor of 0.0005.
To improve training efficiency, a cosine annealing learning rate scheduler as in \eqref{eq-annealing} was employed with a minimum learning rate of $10^{-5}$. 
\begin{equation} 
\eta_t = \eta_{min}+\frac{1}{2}(\eta_{max}-\eta_{min})\left(1+\cos\left(\frac{T_{cur}}{T_{max}}\pi\right)\right),\label{eq-annealing} 
\end{equation} 
where $T_{cur}$ is the current epoch, $T_{max}$ denotes the total number of epochs, and $\eta_t$ is the learning rate at epoch $t$.
The combination of SGD with momentum and cosine annealing effectively balances convergence speed with robust optimization. These configurations achieve stable training and improve the model's generalization performance.

Hence, cross-entropy loss as in \eqref{eq-ce} was employed to guide the model optimization process.
\begin{equation}
l(x,y)=-\sum_{n=1}^{N}log\frac{exp(x_{n,y_n})}{\sum_{c=1}^{C}exp(x_{n,c})}, \label{eq-ce}
\end{equation}
where $x$ is the input, $y$ is the target, $C$ is the number of classes, and $N$ spans the minibatch dimension.

 During training, a batch size of 8 was used, along with data augmentation techniques, viz. random horizontal and vertical flips, 90-degree rotations, and random adjustments to hue, saturation, and value. To prevent overfitting, early stopping was applied with a patience threshold of 30 epochs.

\begin{table*}[!tp]
\setlength{\tabcolsep}{4.5pt} 
\caption{mIoUs and class IoUs of previous semantic segmentation models and our proposed mobilenetv3-large deeplabv3 model on Agriculture-Vision \textbf{validation} set. \texttt{BG} - Background, \texttt{DP} - Double Plant, \texttt{DD} - Dry Down, \texttt{ER} - End Row, \texttt{ND} - Nutrient Deficiency, \texttt{PS} - Planter Skip, \texttt{WA} - Water, \texttt{WW} - Waterway, \texttt{WC} - Weed Cluster; $\uparrow$ - improvement, $\downarrow$ - declination in overall mIoU compared to the baseline}
\label{results-table}
\begin{center}
\begin{small}
\begin{sc}
\begin{tabular}{l|cc|ccccccccc}
\hline
  \multirow{2}*{Method}   & \centering mIoU & \centering gain & \multicolumn{9}{c}{Semantic Class-specific Model Performance in IoU $\%$} \\ \cline{4-12} 
 & \centering $\%$ & \centering $\%$ & \texttt{BG} & \texttt{DP}  & \texttt{DD} & \texttt{ER} & \texttt{ND} & \texttt{PS} & \texttt{WA} & \texttt{WW} & \texttt{WC} \\
\hline
HRNetV2+OCR \cite{tavera2022augmentation} & 34.42 & $\downarrow$20.69 & 72.60 & 17.98 & 56.69 & 11.97 & 27.91 & 23.79 & 48.99 & 27.73 & 22.06 \\
DeepLabv3 (os=8) \cite{chiu2020agriculture}  & 35.29 & $\downarrow$18.69 & 73.01  & 21.32  & 56.19 & 12.00 & 35.22 & 20.10 & 42.19 & 35.04  & 22.51 \\
DeepLabv3+ (os=8) \cite{chiu2020agriculture}  & 37.95 & $\downarrow$12.56 & 72.76          & 21.94          & 56.80        & 16.88          & 34.18          & 18.80          & 61.98          & 35.25          & 22.98 \\
DeepLabv3 (os=16) \cite{chiu2020agriculture}  & 41.66          & $\downarrow$~~4.01 & 74.45          & 25.77          & 57.91        & 19.15          & 39.40          & 24.25          & 72.35          & 36.42          & 25.24 \\
DeepLabv3+ (os=16) \cite{chiu2020agriculture}  & 42.27          & $\downarrow$~~2.60 & 74.32          & 25.62          & 57.96        & 21.65          & 38.42          & 29.22          & 73.19          & 36.92          & 23.16 \\
Agri-vision baseline \cite{chiu2020agriculture} & 43.40 & \centering -- & 74.30 & 28.50 & 57.40 & 21.70 & 38.90 & 33.60 & 73.60 & 34.40 & 28.30 \\
MiT-B3 w/o Boundary \cite{shen2022aaformer}  & 45.31 & $\uparrow$~~4.40 & 77.43 & 37.26 & 59.88 & 24.13 & 42.41 & 41.64 & 69.29 & 26.63 & 29.13 \\
MiT-B3 w/t Boundary \cite{shen2022aaformer}  & 45.44          & $\uparrow$~~4.70 & {76.85} & 37.06          & 60.93        & 24.45          & 42.42          & 41.35          & 69.17          & 26.89          & 29.89 \\
SegFormer \cite{tavera2022augmentation} & 46.50 & $\uparrow$~~7.14 & 76.17 & 33.63 & 58.96 & 18.92 & 40.57 & 38.93 & 80.56 & 42.85 & 27.88 \\
\textbf{Ours} & \textbf{47.17} & $\uparrow$~~\textbf{8.69} & \textbf{74.39} & \textbf{31.34} & \textbf{59.32 }& {\textbf{25.19}} & \textbf{42.67} & {\textbf{43.75}} & \textbf{74.55} & \textbf{39.08} & {\textbf{34.28}} \\
SegFormer+MiT-B5 \cite{tavera2022augmentation} & {49.04} & $\uparrow$12.99 & 76.19 & {37.32} & \text{61.75} & 24.57 & {42.75} & 42.01 & {81.32} & {43.71} & 31.75 \\
\hline
\end{tabular}
\end{sc}
\end{small}
\end{center}
\end{table*}

\begin{table*}[!tp]
\caption{Effectiveness Comparison of the Best Performing Models Compared to the Baseline}\label{tab-model-effectiveness}
\setlength{\tabcolsep}{6.0pt} 
\begin{center}
\begin{small}
\begin{sc}
\begin{tabular}{lccccr}
\toprule
Method & $\text{Diff}_{mIoU}$ & Params (M) & GFLOPs & Effectiveness ($\%$)\\
\midrule
Agri-vision baseline \cite{chiu2020agriculture} & baseline & 60.9 & 258.74 & baseline \\
MiT-B3 w/t Boundary \cite{shen2022aaformer} & 2.04 & 47.3 & ~~79.0 & ~1.54\\
SegFormer \cite{tavera2022augmentation} & 3.10 & 84.7 & ~~~183.3~~ & ~~0.87\\
\textbf{Ours}    & \textbf{3.77} & ~~\textbf{7.6} & ~~~\textbf{6.32} & \textbf{67.77}\\
SegFormer+MiT-B5 \cite{tavera2022augmentation} & 5.64 & 84.7 & ~~~~183.3~~ & ~1.60\\

\bottomrule
\end{tabular}
\end{sc}
\end{small}
\end{center}
\end{table*}

\subsection{Evaluation Metrics}

This study uses the widely accepted intersection over union (IoU) as the evaluation metric. The mean IoU (mIoU) is calculated by averaging the intersection over union for all semantic classes, as follows:
\begin{equation}
mIoU = \frac{1}{k} \sum_{i=1}^{k} \frac{P_{ii}}{\sum_{j=1}^{k} P_{ij} + \sum_{j=1}^{k} P_{ji} - P_{ii}}, 
\end{equation}
where $k$ is the number of semantic classes, $P_{ij}$ denotes the number of pixels where the true label is class $i$ and the predicted label is class $j$, $P_{ji}$ represents the number of pixels, where the true label is class $j$ and the predicted label is class $i$, and $P_{ii}$ is the number of pixels where both the true and predicted labels are class $i$.

The model effectiveness, in \cref{eq-effectiveness}, serves as a unified metric that accounts for both computational complexity and predictive performance: 
\begin{equation} 
    \text{Effectiveness} = \frac{\text{Diff}_{mIoU}}{\log(\text{Params}) \cdot \text{GFLOPs}} \times 100\%, \label{eq-effectiveness} 
\end{equation} 
where $\text{Diff}_{mIoU}$, Params, and GFLOPs represent the mIoU difference from the baseline, the number of parameters, and the computational cost, respectively. The $\log(\text{Params})$ term ensures that models with very high number of parameters do not disproportionately dominate the metric.

\begin{figure*}[!ht]
\vskip 0.2in
\begin{center}
{\includegraphics[trim={0.25cm, 2.45cm, 0cm, 0cm}, clip, width=1\linewidth]{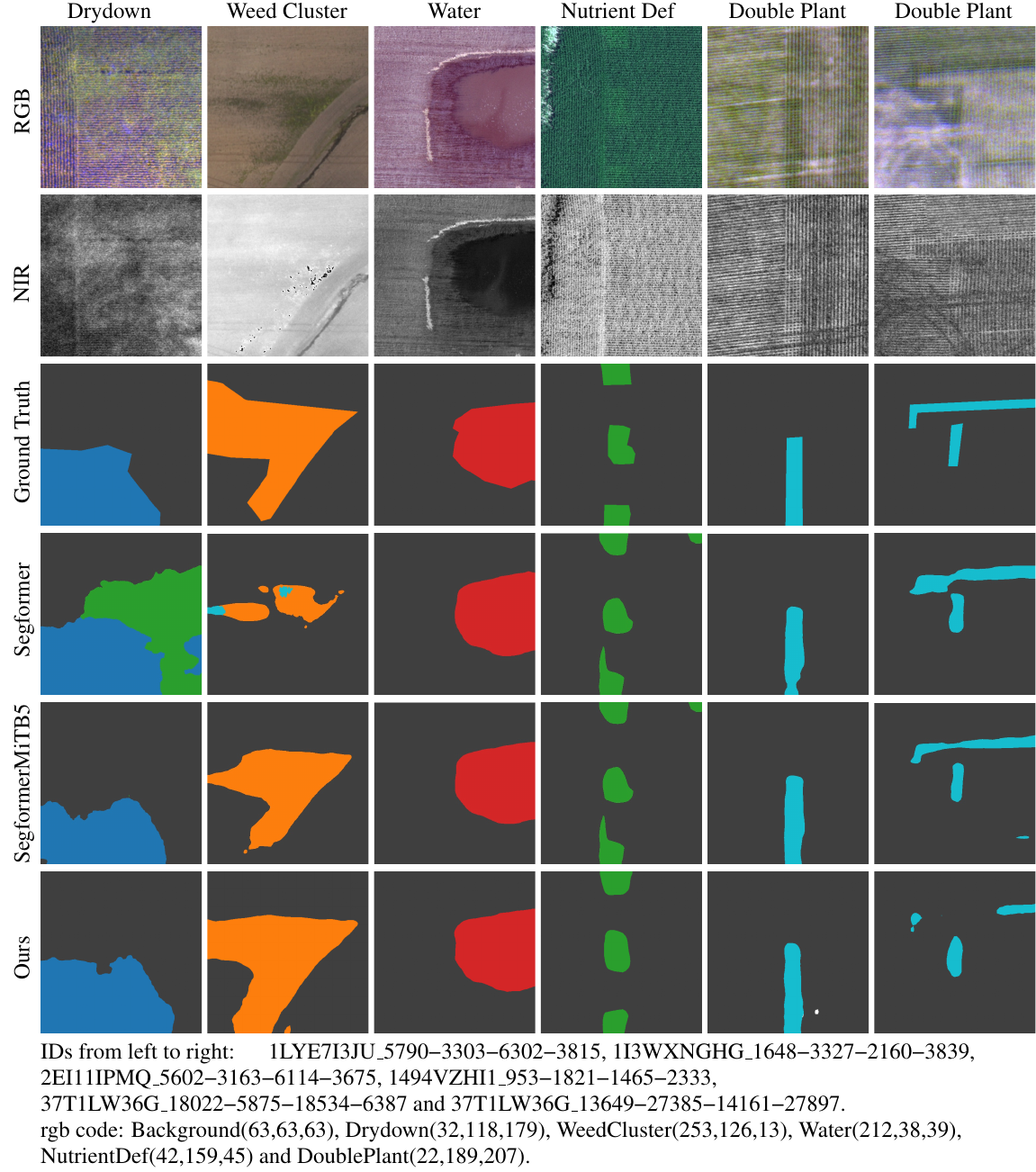} }\\
\vspace{-.2cm}
\begin{flushleft}{\footnotesize The actual file name of the input samples from left to right: \texttt{1LYE7I3JU\char`_5790-3303-6302-3815}, \texttt{1I3WXNGHG\char`_1648-3327-2160-3839}, 
\texttt{2EI11IPMQ\char`_5602-3163-6114-3675}, \texttt{1494VZHI1\char`_953-1821-1465-2333}, \texttt{
37T1LW36G\char`_18022-5875-18534-6387}, \texttt{37T1LW36G\char`_13649-27385-14161-27897.}}\end{flushleft}
\caption{Sample qualitative results. The Segformer and SegformerMiT-B5 results are adopted from Tavera et al.~\cite{tavera2022augmentation}.}
\label{fig-viz-SegMiTB5-tavera}
\end{center}
\end{figure*}

\subsection{Comparative Analysis}

For a comprehensive comparative analysis, this study evaluates related works in the literature that reported a mean Intersection over Union (mIoU) greater than $30\%$, as summarized in \cref{results-table}. Our proposed model achieves a mIoU of $47.17\%$, representing an $8.69\%$ improvement over the baseline model. Furthermore, it demonstrates superior performance in key categories, including Endrow, Planter Skip, and Weed Cluster, when compared to highly complex ViT-based SOTA methods.

\textbf{Computational complexity:} Our model comprises $\approx 7.59$M parameters and requires $6.32$ GFLOPs for inference. By contrast, ViT-based SOTA methods, such as SegFormer~\cite{tavera2022augmentation}, are built on versions of the MiT architecture and exhibit significantly higher complexity. For example, the MiT-B3 and MiT-B5 backbones have $\approx 47.3$M parameters with $79$ GFLOPs and $84.7$M parameters with $183.3$ GFLOPs, respectively~\cite{Gu_2022_CVPR, NEURIPS2021_64f1f27b}. This implies that our model is roughly $6\times$ smaller than MiT-B3 and $11\times$ smaller than MiT-B5 while requiring significantly fewer computational resources.
Besides, our model surmounts the competing approaches in model efficiency, achieving a score of $\approx 67\%$, as analyzed in \cref{tab-model-effectiveness}. This proves its ability to balance predictive performance and computational efficiency, marking a significant advancement in lightweight, yet effective segmentation frameworks.

\Cref{fig-viz-SegMiTB5-tavera} presents qualitative comparisons of the proposed model against the two best-performing models reported by Tavera et al.~\cite{tavera2022augmentation}. For consistency, we use the same validation images as those in Tavera et al.~\cite{tavera2022augmentation}. A closer examination of these qualitative results reveals strong alignment with the quantitative findings in \cref{results-table}, demonstrating that our model outperforms existing methods across all categories except for the second Double Plant. Furthermore, our model achieves a higher mIoU than Segformer+MiT-B5 in the Weed Cluster category, as shown in the figure, which also aligns with the quantitative analysis in \cref{results-table}. Additional qualitative results are provided in \cref{appendix-More quantitative results}.

\section{Conclusion}\label{conclusion}

This work introduces an efficient approach that enhances the Atrous Spatial Pyramid Pooling module of DeepLabV3 by replacing traditional atrous convolutions with the proposed Dual Atrous Separable Convolution dubbed \texttt{DAS-Conv}. The dilation rates and paddings in \texttt{DAS-Conv} are meticulously optimized to improve model performance. Furthermore, through empirical analysis, skip connections are strategically integrated within the Con-EnDec architecture, enhancing the model's ability to capture rich features and spatial hierarchies.
The proposed refinements surpass the baseline and existing models, achieving competitive semantic segmentation results compared to SOTA, while demonstrating higher effectiveness, highlighting its practicality for real-world applications.

Several promising directions for future work are proposed: (i) evaluating the model on diverse datasets to assess its generalizability across different domains, (ii) incorporating attention mechanisms to improve the model's ability to focus on relevant features, thereby enhancing its performance in more complex tasks, and (iii) extending the approach to a semi-supervised learning framework to address the challenge of limited labeled data, further broadening its applicability to real-world scenarios.


\section*{Impact Statement}

This work proposes a highly efficient semantic segmentation model for farmland anomaly detection by integrating a dual atrous separable convolutional module into a convlutional encoder-decoder architecture. The findings of this research have the potential to facilitate large-scale precision agriculture, enhancing food security and environmental sustainability by optimizing land utilization and minimizing resource waste. 




\bibliography{MyPaper}
\bibliographystyle{plain}

\newpage
\appendix
\onecolumn

\section{The DAS-Conv Module}







The code snippet of the proposed \texttt{DAS-Conv} module is given below. 

\begin{figure}[!ht]
\begin{center}
{\includegraphics[trim={2cm, 8.6cm, 2.5cm, 4cm}, clip, width=\columnwidth]{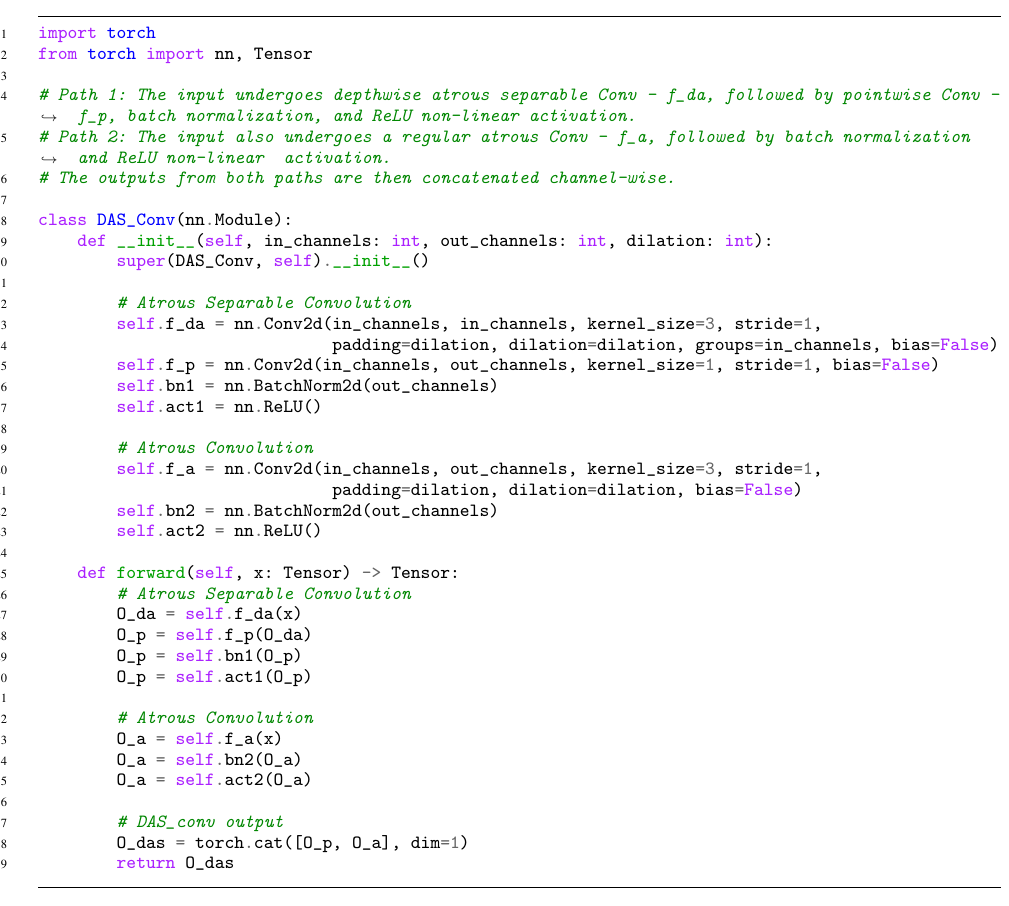} }
\label{fig-das-code}
\end{center}

\end{figure}

\section{Additional Qualitative Results}
\label{appendix-More quantitative results}

\begin{figure*}[ht]
\vskip 0.2in
\begin{center}
{\includegraphics[trim={0.25cm, 1.5cm, 0cm, 0cm}, clip, width=1.0\linewidth]{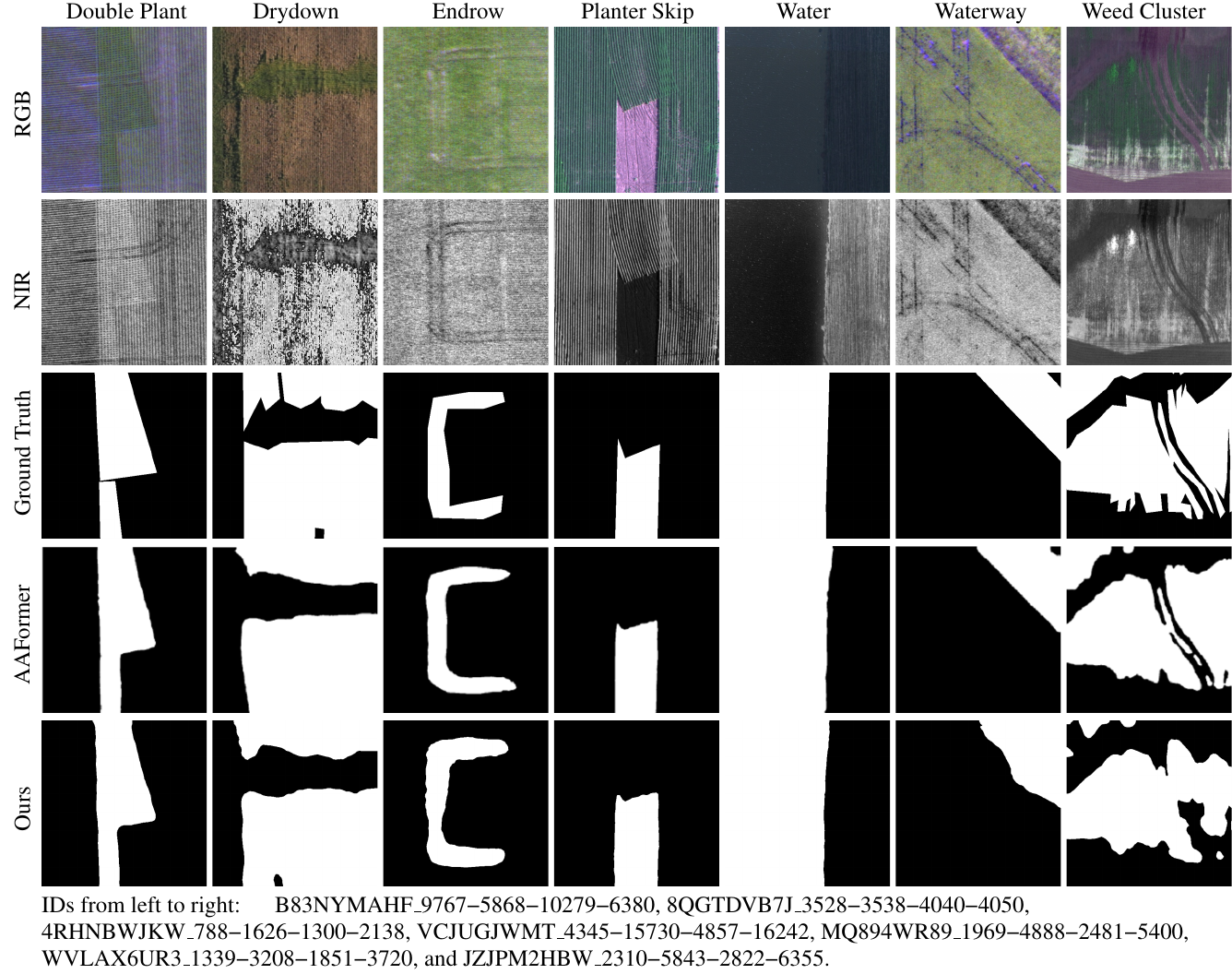} }\\
\vspace{-.2cm}
\begin{flushleft}{\footnotesize The actual file name of the input samples from left to right: \texttt{B83NYMAHF\char`_9767-5868-10279-6380}, \texttt{8QGTDVB7J\char`_3528-3538-4040-4050}, \texttt{4RHNBWJKW\char`_788-1626-1300-2138}, 
\texttt{VCJUGJWMT\char`_4345-15730-4857-16242}, \texttt{ MQ894WR89\char`_1969-4888-2481-5400}, \texttt{WVLAX6UR3\char`_1339-3208-1851-3720}, \texttt{JZJPM2HBW\char`_2310-5843-2822-6355.}}\end{flushleft}
 \caption{Additional qualitative results. The results of AAFormer (MiT-B3 w/t boundary) are adopted from \cite{shen2022aaformer}.}
\label{fig-viz-MiT-B3-boundary-shen}
\end{center}
\vskip -0.2in
\end{figure*}

\cref{fig-viz-MiT-B3-boundary-shen} presents an additional qualitative comparison with the results of Shen~et~al.~\cite{shen2022aaformer}. For consistency, we employ the same validation images used in their study. Upon closer inspection, it is evident that the AAFormer with MiT-B3 achieves slightly better segmentation results in the Waterway and Weed Cluster categories, while our model outperforms theirs in all categories except for Background, Double Plant, and Drydown. This observation aligns with the quantitative results presented in \cref{results-table}.

\begin{figure*}[ht]
\vskip 0.2in
\begin{center}
{\includegraphics[trim={0.25cm, 0cm, 0cm, 0cm}, clip, width=1.0\linewidth]{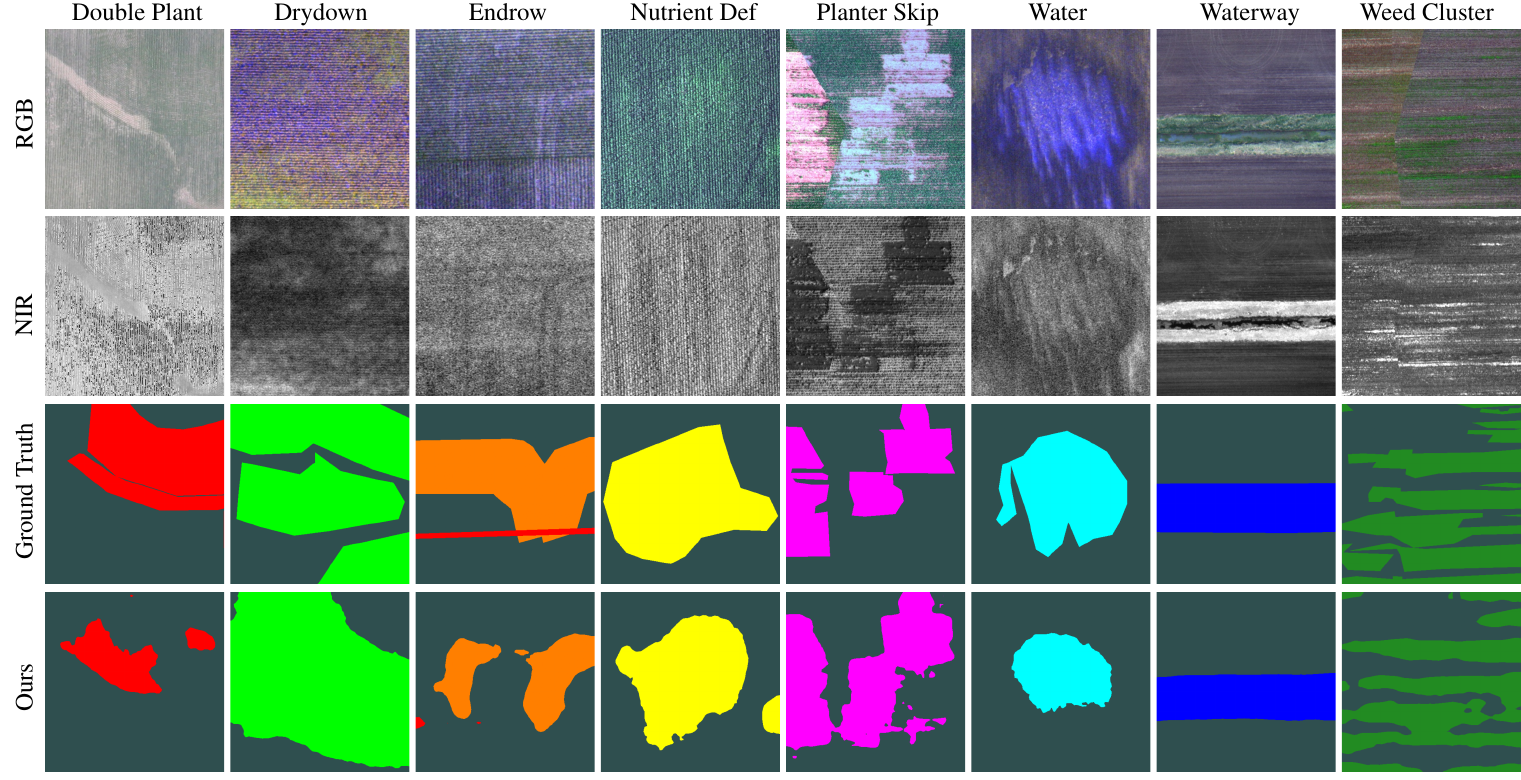} }\\
\vspace{-.2cm}
\begin{flushleft}
{\footnotesize The actual file name of the input samples from left to right: 
\texttt{PRGNELPEZ\char`5711-6690-6223-7202}, \texttt{1LYE7I3JU\char`7768-2989-8280-3501},
\texttt{FAKX1V3ZX\char`3621-664-4133-1176}, \texttt{QVEJC9GFZ\char`3809-9076-4321-9588},
\texttt{2PTXX7UQM\char`9178-5979-9690-6491}, \texttt{MUPQIEZP4\char`2989-2040-3501-2552},
\texttt{G2APW2Z2Z\char`4307-4204-4819-4716}, \texttt{QEQNXLHLY\char`3250-3118-3762-3630}.
}
\end{flushleft} \vspace{-.2cm}
 \caption{Qualitative samples illustrating the proposed model's failure case compared to the ground truth.}
\label{fig-viz-our_model}
\end{center}
\vskip -0.2in
\end{figure*}

\subsection{Failure Analysis} 

In addition to the previous qualitative comparisons between the proposed models and SOTA models, \cref{fig-viz-our_model} presents a standalone visualization of our model’s performance, specifically to analyze its failure cases. The model exhibits challenges in accurately segmenting the Double Plant and Endrow categories while demonstrating satisfactory performance across the remaining classes. This analysis highlights areas for potential improvement, suggesting that further refinements could enhance the model’s overall effectiveness.

\section{Model Building Process and Ablation Studies}

\begin{table}[!htp]
\caption{Summary of the model building exploration and ablation studies. Note: OPT - optimizer, Skip - the application skip connection from last encoder's module with stated dimension of the feature map to the decoder. $f_{\beta}\{d_1, \cdots,d_n\}$ refers to the specific atrous function type $\beta$ with a $n$ number of various dilation values of $d_1$, $d_2$, $\cdots$, and $d_n$. For example, $f_{das}\{4,8,12\}$ means that besides the Conv $1\times1$ and global average pool, the ASPP module also consists of three \texttt{DAS-Conv} with dilation rates of 4, 8 and 12, respectively. } \label{table-ablation-studies}
\begin{center}
\begin{small}
\setlength{\tabcolsep}{5pt} 
\renewcommand{\arraystretch}{1.1} 
\begin{tabular}{p{0.05cm} p{0.05cm} p{3.3cm} c c c c c c p{2.2cm} c}
\toprule
\multirow{19}{*}{\rotatebox{90}{\sc Model Building Exploration Stages}}  &  &\sc Model & \sc Input & \sc Aug & \sc Opt & \sc GFLOPs & \sc Param & \sc Skip & 
\sc $f_{\beta}\{d_1, \cdots,d_n\}$ & \sc mIoU, $\%$\\
\midrule \midrule
&\multirow{2}{*}{1} & Resnet50 Deeplabv3 & RGB & No & Adam & 163.57 & 42M & No & $f_a\{12,24,32\}$ & 31.58\\ 
& & Resnet50 Deeplabv3 & RGB & No & SGD & 163.57 & 42M & No & $f_a\{12,24,32\}$ & 35.75\\ 

\cline{2-11}
& \multirow{2}{*}{2} & Resnet50 Deeplabv3 & RGBn & No & Adam & 163.77 & 42M & No & $f_a\{12,24,32\}$ & 33.46\\
& & Resnet50 Deeplabv3 & RGBn & No & SGD & 163.77 & 42M & No & $f_a\{12,24,32\}$ & 37.71\\
\cline{2-11}

& \multirow{5}{*}{3} & Resnet50 Deeplabv3 & nRGB & Yes & SGD & 163.77 & 42M & No & $f_a\{12,24,32\}$ & 37.35\\
& & Resnet101 Deeplabv3 & nRGB & No & SGD & 241.25 & 61M & No & $f_a\{12,24,32\}$ & 38.66\\
& & Resnet101 Deeplabv3 & nRGB & Yes & SGD & 241.25 & 61M & No & $f_a\{12,24,32\}$ & 42.27\\
& & Mobilenetv3L Deeplabv3 & nRGB & No & SGD & 9.83 & 11M & No & $f_a\{12,24,32\}$ & 38.59\\
& & Mobilenetv3L Deeplabv3 & nRGB & Yes & SGD & 9.83 & 11M & No & $f_a\{12,24,32\}$ & 41.70\\

\cline{2-11}
& \multirow{3}{*}{4} & Mobilenetv3L Deeplabv3 & nRGB & Yes & SGD & 3.29 & 4.6M & $32^2\times960$ & $f_{da}\{12,24,32\}$ & 41.66\\
& & Mobilenetv3L Deeplabv3 & nRGB & Yes & SGD & 3.29 & 4.6M & $64^2\times40$ & $f_{da}\{12,24,32\}$ & 42.45\\
& & Mobilenetv3L Deeplabv3 & nRGB & Yes & SGD & 3.32 & 4.6M & $128^2\times24$ & $f_{da}\{12,24,32\}$ & 41.38\\
\cline{2-11}
& \multirow{5}{*}{5} & Mobilenetv3L Deeplabv3 & nRGB & Yes & SGD & 3.29 & 4.6M & $64^2\times40$ & $f_{da}\{12,24,32\}$ & 45.46\\
& & Mobilenetv3L Deeplabv3 & nRGB & Yes & SGD & 6.31 & 7.5M & $64^2\times40$ & $f_{das}\{12,24,32\}$ & 46.40\\
& & Mobilenetv3L Deeplabv3 & nRGB & Yes & SGD & 6.31 & 7.5M & $64^2\times40$ & $f_{das}\{4,8,12\}$ & 46.61\\
& & Mobilenetv3L Deeplabv3 & nRGB & Yes & SGD & 6.32 & 7.6M & $64^2\times40$ & $f_{das}\{4,8,12,24\}$ & 47.17\\
& & Mobilenetv3L Deeplabv3 & nRGB & Yes & SGD & 6.35 & 7.6M & $128^2\times24$ & $f_{das}\{4,8,12,24\}$ & 46.65\\

\bottomrule
\end{tabular}
\end{small}
\end{center}
\vskip -0.1in
\end{table}

To derive the best-performing model, we adopted a bottom-up approach, starting with a vanilla Conv-EnDec model using ResNet50 with DeepLabV3. We then conducted a series of ablation studies to evaluate various configurations, including different forms of atrous convolution: standard atrous convolution, $f_a(\cdot)$, depth-wise atrous convolution, 
$f_{da}(\cdot)$, and depth-wise separable atrous convolution, $f_{das}(\cdot)$. Additionally, we explored the impact of different dilation rates and other hyperparameter variations.

Given the large scale and high resolution of the Agriculture-Vision dataset~\cite{chiu2020agriculture}, conducting multiple ablation studies would result in prolonged training times. To address this, we created a Mini Aggri-Vision datset, which is a smaller subset of the original dataset, for preliminary experimentation by randomly sampling 1/8 of the entire training and validation sets. This subset allowed efficient model exploration while maintaining a representative distribution of the original dataset. Although we conducted hundreds of experiments, for clarity and structured presentation, we consolidate our key findings into five stages of exploration to find the best optimizer (Adam vs SGD), the best multi-spectral input channel arrangement (nRGB vs RGBn), the best encoder backbone, the best skip connection source, and the best separable Convs with good dilation rates. It comprises a total of seventeen different experiments, as summarized in \Cref{table-ablation-studies}.

\subsection{Stage 1 Exploration: Quest for the Best Optimizer}

Using the mini dataset, we first implemented the vanilla model with RGB images as input. We then assessed the performance of different optimizers, specifically Adam and SGD, as presented in \Cref{table-ablation-studies}. The results indicate that the SGD optimizer achieves a higher mIoU, suggesting its suitability for our model configuration in this case.

\subsection{Stage 2 Exploration: Quest for the Best Input Data Type}

With the availability of the Near-Infrared (NIR) channel, we extended our experiments to include RGBn as the model input while maintaining the same configurations as in Stage 1. 
The results reaffirm that the SGD optimizer consistently outperforms Adam in this setup. Furthermore, the findings indicate that the model that uses the 4-channel RGBn (multi-spectral data) input achieves better performance compared to the standard 3-channel RGB model, highlighting the advantage of incorporating NIR information for improved segmentation accuracy.

\subsection{Stage 3 Exploration: Quest for the Best Multi-spectral Input Channel Arrangement and Encoder Backbone}

At this stage, we explored an alternative 4-channel input arrangement, nRGB, and evaluated other widely used backbone architectures, including ResNet101 and MobileNetV3Large, as the vision encoder for DeepLabV3. Data augmentation techniques were also implemented at this stage and thus, produced higher performance than those of experiments without them. The results also indicate that both DeepLabV3 with ResNet101 and MobileNetV3Large outperformed the ResNet50 variant in terms of mIoU. 
Although DeepLabV3 with ResNet101 achieves the highest performance, we opted to proceed with DeepLabV3 using MobileNetV3Large due to its significantly smaller model size-more than $15\times$ smaller than the other architectures-while maintaining competitive performance. This lightweight model is particularly well-suited for this study.


\subsection{Stage 4 Exploration: Quest for the Optimal Placement of a Skip Connection}

In Stage 4, we substituted all atrous Conv in the ASSP module with atrous separable Conv with the same dilation rates and introduced a skip connection from one of the last modules of the vision encoder with different spatial dimensions. Specifically, the encoder contains three key points where low-level feature maps are extracted: $32\times32\times960$, $64\times64\times40$, and $128\times128\times24$. 
A stacking point is established in the decoder, where these low-level feature maps from the encoder are combined via the skip connection and concatenated channel-wise with the high-level feature maps, as illustrated in \cref{overall architecture}.

In addition, the architecture incorporates two upsampling stages in the decoder. The first upsampling factor corresponds to the ratio between the spatial dimensions of the feature maps transferred through the skip connection and those output by the ASPP module. The second upsampling factor is determined by the ratio of the desired segmentation map resolution to the spatial dimensions of the stacked feature maps. For example, if the skip connection transfers a $32\times32\times960$ feature map, the corresponding upsampling factors would be 1 and 16, respectively. Similarly, if the feature map of the skip connection is $128\times128\times24$, the corresponding upsampling factors would be 4 and 4. The experimental results indicate that using the $64\times64\times40$ feature maps produces the highest mIoU. 

At this stage, we also performed numerous additional tests, exploring various encoders, including different sections of the Inception network, as well as alternative decoder architectures, such as DenseASPP. Furthermore, various modifications to the MobileNetv3 DeepLabv3 architecture were employed, and adaptive weight techniques to loss function were experimented. 
However, none of these configurations exceeded an mIoU of 42\%, suggesting that the model may require a larger volume of training data to generalize better. 

\subsection{Stage 5 Exploration: Quest for the Best Combination of Dilation Rates for Our Novel \texttt{DAS-Conv} Module}

In this stage, we used the entire samples (56,944 images) of the original Agriculture-Vision dataset, which led to an improved model performance when applying the best performing configuration of using a skip connection with a dimension feature map $64\times64\times40$ and nRGB as input to the model from Stage 4.

In previous stages, we employed either atrous separable Conv or atrous Conv individually but never in combination. Given that both techniques capture feature maps differently, we experimented with their simultaneous use, which resulted in an increase in mIoU. Thus, we named this new combination of both Conv types as Dual Atrous Separable Conv, abbreviated as \texttt{DAS-Conv}.

Next, we observed that the originally used dilation rates of $\{12, 24, 32\}$ in the atrous and atrous separable convolutions might be too large for the $32\times32\times960$ feature maps from the encoder backbone entering the ASPP module. To address this,  smaller dilation rates of $\{4, 8, 12\}$ were implemented, which further improved performance. Building on this, we introduced another dilation rate of 24 to the current configuration, making the combination of dilation rates to be $\{4, 8, 12, 24\}$ leading to an additional boost in mIoU. Thus, this is the best model configuration that this study focuses on. 

Lastly, we explored modifying the skip connection in the encoder to transfer higher-resolution feature maps ($128\times128\times24$). However, this adjustment led to a decline in performance, consistent with the findings in Stage 4. Consequently, we did not pursue further experiments in this direction for this study.

In summary, the optimal configurations identified from the model-building ablation studies are as follows: (a) Encoder backbone – MobileNetV3-Large, (b) Input data – multi-spectral (nRGB), (c) Decoder – an enhanced DeepLabV3 with optimized skip connections transferring $64\times64\times40$ feature maps from the encoder and our \texttt{DAS-Conv} module with dilation rates of ${4,8,12,24}$, and (d) Optimizer – SGD.

\end{document}